\documentclass{article} 

\usepackage[dvipsnames,table,xcdraw]{xcolor}

\usepackage{iclr2022_conference,times}

\usepackage{hyperref}
\usepackage{url}

\usepackage{multirow}
\usepackage[detect-all]{siunitx}

\usepackage{pgfplots}

\usepackage{amsfonts}
\usepackage{amsmath}
\usepackage{amssymb}
\usepackage[inline]{enumitem}
\usepackage{amsthm}
\usepackage{bm}
\usepackage{allrunes}
\usepackage{etoolbox}
\usepackage{xspace}
\usepackage{appendix}

\usepackage{algorithm2e}

\usepackage[textsize=footnotesize]{todonotes}

\usepackage{tikz}
\usetikzlibrary{automata, positioning, arrows, matrix}

\usepackage{booktabs}

\DeclareMathOperator*{\argmin}{arg\,min}

\renewcommand{\x}{\bm{x}}
\renewcommand{\y}{\bm{y}}
\newcommand{\thetac}{\theta_c}
\newcommand{\thetau}{\theta_u}

\newcommand{\Lang}{\mathcal{L}}
\newcommand{\biLang}{\ddot{\mathcal{L}}}

\newcommand{\biM}{\ddot{M}}

\newcommand{\Pu}[1][\theta]{P_{#1}(\y\mid\x)}
\newcommand{\Pc}[1][\theta]{P_{#1}(\y\mid\x,\Lang)}

\newcommand{\E}[1]{E_{\x, \y \sim \widetilde{P}} \left[#1\right]}

\newcommand{\labeledarrow}[1]{\raisebox{-3pt}{$\xrightarrow{#1}$}}

\newcommand{\edge}[3]{\left(#1\labeledarrow{\text{#2}}#3\right)}

\newcommand{\ct}{\textit{ConstrTr}\xspace}
\newcommand{\cd}{\textit{ConstrDec}\xspace}

\newcommand{\full}{\textit{CRF-full}\xspace}
\newcommand{\red}{\textit{CRF-reduced}\xspace}

\title{Constraining Linear-chain CRFs to Regular Languages}

\author{%
  Sean Papay, Roman Klinger, \& Sebastian Pad\'o\\
  University of Stuttgart\\
  \texttt{(sean.papay|klinger|pado)@ims.uni-stuttgart.de}
}

\iclrfinalcopy 
\newtheorem{theorem}{Theorem}
\begin{document}

\raisebox{1cm}[0pt]{\parbox{\textwidth}{\textcolor{red}{\footnotesize Due to a software bug, some of the evaluation scores we reported in our
original publication were lower than their true values.
These have been corrected, and our analysis was updated to reflect the
true values.}}} \vspace{-1.16cm}\\
\maketitle

\begin{abstract}
  A major challenge in structured prediction is to represent the
  interdependencies within output structures.  When outputs are
  structured as sequences, linear-chain conditional random fields
  (CRFs) are a widely used model class which can learn \textit{local}
  dependencies in the output. However, the CRF's Markov assumption
  makes it impossible for CRFs to represent distributions with
  \textit{nonlocal} dependencies, and standard CRFs are unable to
  respect nonlocal constraints of the data (such as global arity
  constraints on output labels).  We present a generalization of CRFs
  that can enforce a broad class of constraints, including nonlocal
  ones, by specifying the space of possible output structures as a
  regular language $\mathcal{L}$.  The resulting regular-constrained
  CRF (RegCCRF) has the same formal properties as a standard CRF, but
  assigns zero probability to all label sequences not in
  $\mathcal{L}$.  Notably, RegCCRFs can incorporate their constraints
  during training, while related models only enforce constraints
  during decoding.  We prove that constrained training is never worse
  than constrained decoding, and show empirically that it can be
  substantially better in practice. Finally, we demonstrate a
  practical benefit on downstream tasks by incorporating a RegCCRF
  into a deep neural model for semantic role labeling, exceeding
  state-of-the-art results on a standard dataset.
\end{abstract}

\linepenalty=1000

\section{Introduction}

\par Structured prediction is a field of machine learning where
outputs are expected to obey some predefined discrete
structure. Instances of structured prediction with various output
structures occur in many applications, including computer vision
(e.g., scene graph generation \citep{johnson2015image} with
graph-structured output), natural language processing (e.g.,
linguistic parsing \citep{niculae2018sparsemap} with tree-structured
output, relation extraction \citep{roth2004linear} with
tuple-structured output) or modeling the spatial structure of physical
entities and processes \citep{9006803}.

A key difficulty faced by all models is to tractably model
interdependencies between different parts of the output.  As output
spaces tend to be combinatorially large, special techniques,
approximations, and independence assumptions must be used to work with
the associated probability distributions. Considerable research has
investigated specific structures for which such approaches make
machine learning tractable.  For instance, when outputs are trees over
a fixed set of nodes, maximal arborescence algorithms allow for exact
inference \citep{chu1965shortest, branchings}; when outputs are
graph-structured, loopy belief propagation can provide approximate
inference \citep{loopy}.

If the output forms a linear sequence, a Markov assumption is often
made: model outputs depend only on their immediate neighbors, but not
(directly) on more distant ones.  A common model that uses this
assumption is the linear-chain conditional random field (CRF)
\citep{lafferty2001conditional}, which has found ubiquitous use for
sequence labeling tasks, including part-of-speech tagging
\citep{pos-twitter} and named entity recognition
\citep{lample-etal-2016-neural}. This model became popular with the
use of hand-crafted feature vectors in the 2000s, and is nowadays
commonly used as an output layer in neural networks to encourage the
learning of structural properties of the output sequence. The Markov
assumption makes training tractable, but also limits the CRF's
expressive power, which can hamper performance, especially for long
sequences \citep{scheible2016model}.  Semi-Markov CRFs
\citep{sarawagi2004semi} and skip-chain CRFs
\citep{sutton2004collective} are techniques for relaxing the Markov
assumption, but both come with drawbacks in performance and
expressiveness.

In this work, we propose a new method to tractably relax the Markov
assumption in CRFs.
Specifically, we show how to constrain the output of a CRF to
\textit{a regular language}, such that the resulting
\textit{regular-constrained CRF (RegCCRF)} is guaranteed to output
label sequences from that language. Since regular languages can encode
long-distance dependencies between the symbols in their strings,
RegCCRFs provide a simple model for structured prediction guaranteed
to respect these constraints. The domain knowledge specifying these
constraints is defined via regular expressions, a straightforward,
well understood formalism.
We show that our method is distinct from approaches that enforce
constraints at decoding time, and that it better approximates the true
data distribution. We evaluate our model empirically as the output
layer of a neural network and attain state-of-the-art performance for
semantic role labeling \citep{ontonotes, pradhan-etal-2012-conll}. Our
PyTorch RegCCRF implementation can be used as a drop-in replacement
for standard CRFs.

\section{Related work}
\label{sec:related-work}

We identify three areas of structured prediction that are relevant to our work:
constrained decoding, which can enforce output constraints at decoding
time, techniques for weakening the Markov assumption of CRFs to learn
long-distance dependencies, and weight-learning in finite-state
transducers.

\paragraph{Constrained decoding.}
A common approach to enforcing constraints in model output is
\textit{constrained decoding}: Models are trained in a standard
fashion, and decoding ensures that the model output satisfies the
constraints. This almost always corresponds to finding or
approximating a version of the model's distribution conditionalized on
the output obeying the specified constraints. This approach is useful
if constraints are not available at training time, such as in the
interactive information extraction task of
\citet{kristjansson2004interactive}. They present \textit{constrained
  conditional random fields}, which can enforce that particular tokens
are or are not assigned particular labels (positive and negative
constraints, respectively).
Formally, our work is a strict generalization of this approach, as
position-wise constraints can be formulated as a regular language, but
regular languages go beyond position-wise constraints. Other studies
treat decoding with constraints as a search problem, searching for the
most probable decoding path which satisfies all constraints.  For
example, \cite{he2017deep} train a neural network to predict
token-wise output probabilities for semantic role labeling following
the BIO label-alphabet \citep{ramshaw1999text}, and then use exact A*
search to ensure that the output forms a valid BIO sequence and that
particular task-specific constraints are satisfied.
For autoregressive models, constrained beam search
\citep{hokamp2017lexically, anderson-etal-2017-guided,
  hasler-etal-2018-neural} can enforce regular-language constraints
during search.
We further discuss constrained decoding as it relates to RegCCRFs in
Section~\ref{sec:constrained_decoding}.

\paragraph{Markov relaxations.}
While our approach can relax the Markov assumption of CRFs 
through nonlocal hard constraints,
another strand of work has developed models which can directly
\textit{learn} nonlocal dependencies in CRFs:
\textit{Semi-Markov CRFs} \citep{sarawagi2004semi} relax the Markov
property to the semi-Markov property. In this setting, CRFs are tasked
with segmentation, where individual segments may depend only on their
immediate neighbors, but model behavior within a particular segment
need not be Markovian. As such, semi-Markov CRFs are capable of
capturing nonlocal dependencies between output variables, but only to
a range of one segment and inside of a segment.
\textit{Skip-chain CRFs} \citep{sutton2004collective} change the
expressiveness of CRFs by relaxing the assumption that only the linear
structure of the input matters: they add explicit dependencies between
distant nodes in an otherwise linear-chain CRF. These dependencies are
picked based on particular properties, e.g., input variables of the
same value or which share other properties. In doing so, they add
loops to the model's factor graph, which makes exact training and
inference intractable, and leads to the use of approximation
techniques such as loopy belief propagation and Gibbs sampling.

\paragraph{Weight learning for finite-state transducers.}
While our approach focuses on the task of constraining the CRF
distribution to a known regular language, a related task is that of
learning a weighted regular language from data. This task is usually
formalized as learning the weights of a weighted finite-state
transducer (FST), as in e.g.\ \cite{eisner2002parameter} with directly
parameterized weights and \cite{rastogi-etal-2016-weighting} with
weights parameterized by a neural network. Despite the difference in
task-setting, this task is quite similar to ours in the formal sense,
and in fact our proposal can be viewed as a particularly well-behaved
special case of FST weight learning for an appropriately chosen
transducer architecture and parameterization. We discuss this
connection further in Section~\ref{sec:fsts}.

\section{Preliminaries and notation}
\label{sec:prelim}
As our construction involves finite-state automata and conditional random
fields, we define these here and specify the notation we will use in
the remainder of this work.

\paragraph{Finite-state automata.}
All automata are taken to be nondeterministic finite-state automata (NFAs) without epsilon transitions.
Let such an NFA be defined as a 5-tuple $(\Sigma, Q, q_1, F, E)$,
where
\begin{itemize*}[label={}] 
\item $\Sigma = \{a_1, a_2, ..., a_{|\Sigma|}\}$ is a finite alphabet
  of symbols,
\item $Q = \{q_1, q_2, ..., q_{|Q|}\}$ is a finite set of states,
\item $q_1 \in Q$ is the sole starting state,
  \item $F\subseteq Q$ is a set of accepting states,
  \item and $E \subseteq Q \times \Sigma \times Q$ is a set of directed,
    symbol-labeled edges between states.  The edges define the NFA's
    transition function $\Delta: Q \times \Sigma \rightarrow 2^Q$,
    with $r \in \Delta(q, a) \leftrightarrow (q, a, r) \in E$.
\end{itemize*}
An automaton is said to accept a string $\bm{s} \in \Sigma^*$ iff there exists a contiguous path of edges from $q_1$ to some
accepting state whose edge labels are exactly the symbols of $\bm{s}$.
The \textit{language} defined by an automaton is the set of all such accepted strings.
A language is \textit{regular} if and only if it is the language of some NFA.

\paragraph{Linear-chain conditional random fields.}
Linear-chain conditional random fields (CRFs)
\cite{lafferty2001conditional} are a sequence labeling model.
Parameterized by $\theta$, they use a global exponential model to
represent the conditional distribution over label sequences
$\y = \langle y_1, y_2, ..., y_t \rangle$ conditioned on input
sequences $\x = \langle x_1, x_2, ..., x_t \rangle$:
$\Pu \propto \exp \sum_j f_{\theta}^j(\x, \y)$, with individual
observations $x_i$ coming from some observation space $X$, and outputs
$y_i$ coming from some finite alphabet $Y$. In this work, we use CRFs
for sequence labeling problems, but the dataset labels do not
correspond directly to the CRF's outputs $y_i$.  In order to avoid
ambiguity, and since the term ``state'' already has a meaning for
NFAs, we call $\y$ the CRF's \textit{tag sequence}, and each $y_i$ a
\textit{tag}. The terms \textit{label sequence} and \textit{label}
will thus unambiguously refer to the original dataset labels.

Each $f_\theta^j$ is a potential function of $\x$ and $\y$,
parameterized by $\theta$.  Importantly, in a linear-chain CRF, these
potential functions are limited to two kinds: The \textit{transition
  function} $g_{\theta}(y_i, y_{i+1})$ assigns a potential to each
pair $(y_i, y_{i+1})$ of adjacent tags in $\y$, and the
\textit{emission function} $h_{\theta}(y_i\mid\x, i)$ assigns a
potential to each possible output tag $y_i$ given the observation
sequence $\x$ and its position $i$.
With these, the distribution defined by a CRF is
\begin{equation}
  \Pu \propto \exp \left(\sum_{i=1}^{t-1} g_{\theta}(y_i, y_{i+1}) + \sum_{i=1}^{t} h_{\theta}(\x,y_i,i)\right).
\end{equation}

Limiting our potential functions in this way imposes a Markov
assumption on our model, as potential functions can only depend on a
single tag or a single pair of adjacent tags.  This makes learning and
inference tractable: the forward algorithm \citep{jurafskymartin} can
calculate negative log-likelihood (NLL) loss during training, and the
Viterbi algorithm \citep{viterbi, jurafskymartin} can be used for
inference.  Both are linear in $t$, and quadratic in $|Y|$ in both
time and space.

\section{Regular-constrained CRFs}

Given a regular language $\Lang$, we would like to
constrain a
CRF to $\Lang$.  We formalize this notion of constraint with
conditional probabilities -- a CRF constrained to $\Lang$ is described
by a (further) conditionalized version of that CRF's distribution $\Pu$, conditioned
on the event that the tag  sequence $\y$ is in $\Lang$. 
We write this distribution as
\begin{equation}
\Pc = \begin{cases}
\alpha \cdot \Pu & \text{if } \y \in \Lang \\
0 & \text{otherwise},\\
\end{cases}
\end{equation}
with $\alpha \geq 1$ defined as $\alpha^{-1} = \sum_{\y \in \Lang} \Pu$.

In order to utilize this distribution for machine learning, we
need to be able to compute NLL losses and perform MAP
inference.  As discussed in Section~\ref{sec:prelim}, both of these are
efficiently computable for CRFs.  Thus, if we can construct a separate
CRF whose output distribution can be interpreted as
$P(\y\mid\x, \Lang)$, both of these operations will be available.  We
do this in the next section.

\subsection{Construction}

\begin{figure}
{
\centering
\resizebox{\textwidth}{!}{
\begin{tikzpicture}[->,>=stealth',shorten >=1pt,auto,node distance=2.8cm,
                    semithick]
  
  \tikzstyle{inpath}=[draw=ForestGreen, ultra thick, fill=ForestGreen]

  \node[initial,accepting,state] at (0, 0) (q1) {$q_1$};
  \node[state] at (3, 0) (q2)  {$q_2$};
  \node[accepting,state] at (0, -2) (q3)  {$q_3$};
  \node[state] at (3, -2) (q4)  {$q_4$};
  
  \path (q1)  edge [loop above] node {O} (q1);
  \path[inpath, preaction={draw,line width=1mm}] (q1)  edge  node {\textcolor{ForestGreen}{\textbf{B}}} (q2);
  \path (q2)  edge [loop above] node {I} (q2);
  \path[inpath] (q2) edge node {\textcolor{ForestGreen}{\textbf{O}}} (q4);
  \path (q4)  edge [loop below] node {O} (q4);
  \path[inpath] (q4)  edge node {\textcolor{ForestGreen}{\textbf{B}}} (q3);
  \path (q2)  edge [bend left=15] node {B} (q3);
  \path (q3)  edge [bend left=15] node {B} (q2);
  \path[inpath] (q3)  edge [loop below] node {\textcolor{ForestGreen}{\textbf{I}}} (q3);
  \path (q3)  edge node {O} (q1);

  \draw [-] (5,.5) -- (5,-2.5);

\end{tikzpicture}
\hspace{1cm}
\begin{tikzpicture}
\tikzstyle{var}=[draw, circle, minimum size=1cm, inner sep=0cm]
\tikzstyle{given}=[fill=lightgray]
\tikzstyle{factor}=[draw, rectangle, fill=white]

\node[var] at (0, 0) (y1) {$\edge{q_1}{B}{q_2}$};
\node[var] at (4, 0) (y2) {$\edge{q_2}{O}{q_4}$};
\node[var] at (8, 0) (y3) {$\edge{q_4}{B}{q_3}$};
\node[var] at (12, 0) (y4) {$\edge{q_3}{I}{q_3}$};
\node[var, given] at (6, 3) (x) {$\x$};
\draw (x) -- (y1) node [midway, factor] {$h_\theta(\x,\text{B}, 1)$};
\draw (x) -- (y2) node [midway, factor] {$h_\theta(\x,\text{O}, 2)$};
\draw (x) -- (y3) node [midway, factor] {$h_\theta(\x,\text{B}, 3)$};
\draw (x) -- (y4) node [midway, factor] {$h_\theta(\x,\text{I}, 4)$};
\draw (y1) -- (y2) node [midway, factor] {$g_\theta(\text{B}, \text{O})$};
\draw (y2) -- (y3) node [midway, factor] {$g_\theta(\text{O}, \text{B})$};
\draw (y3) -- (y4) node [midway, factor] {$g_\theta(\text{B}, \text{I})$};
\end{tikzpicture}
}}
\vspace{.15cm}\\
\resizebox{\textwidth}{!}{
$Y' = \left\{\edge{q_1}{O}{q_1}, \edge{q_1}{B}{q_2}, \edge{q_2}{I}{q_2}, \edge{q_2}{B}{q_3}, \edge{q_2}{O}{q_4}, \edge{q_3}{O}{q_1}, \edge{q_3}{B}{q_2}, \edge{q_3}{I}{q_3}, \edge{q_4}{B}{q_3}, \edge{q_4}{O}{q_4}\right\}$
}
\caption{Example for a RegCCRF, showing NFA and unrolled factor
  graph. $\Lang$ describes the language
  $(\text{O}\mid\text{BI}^*\text{O}^*\text{BI}^*)^*$, the language of
  valid BIO sequences for an even number of spans.  We would like to
  calculate $\Pc$ for
  $\y = \langle \text{B}, \text{O}, \text{B}, \text{I} \rangle$.  We
  show an unambiguous automaton $M$ for $\mathcal{L}$ (left), and a
  factor graph (right) for the auxiliary CRF computing
  $P_\theta(\bm{y'}\mid\x)$, where $\bm{y'} \in Y'^*$ corresponds to
  the sole accepting path of $\y$ through $M$ (marked).}
\label{fig:construction}
\end{figure}

Let $M := (\Sigma, Q, q_i, F, E)$ be an NFA that describes $\Lang$.  We
assume that $M$ is \textit{unambiguous} -- i.e., every
string in $\Lang$ is accepted by exactly one path through $M$.  As
every NFA can be transformed into an equivalent unambiguous NFA
\citep{mohri2012disambiguation}, this assumption involves no
loss of generality.
Our plan is to represent $\Pc$ by constructing a separate CRF with a
distinct tag set, whose output sequences can be interpreted directly
as paths through $M$. As $M$ is unambiguous, each label sequence in
$\Lang$ corresponds to exactly one such path.  We parameterize this
auxiliary CRF identically to our original CRF -- that is, with
label-wise (not tag-wise) transition and emission functions. Thus, for
all parameterizations $\theta$, both distributions $\Pu$ and $\Pc$ are
well defined.

There are many ways to construct such a CRF.  As CRF training and
inference are quadratic in the size of the tag set $Y$, we would
prefer a construction which minimizes $|Y|$.  However, for clarity, we
will first present a conceptually simple construction, and discuss
approaches to reduce $|Y|$ in Section~\ref{sec:tagmin}.
We start with our original CRF, parameterized by $\theta$, with tag
set $Y = \Sigma$, transition function $g_\theta$, and emission
function $h_\theta$, describing the probability distribution
$\Pu$, $\y \in \Sigma^*$.  From this, we construct a new CRF, also
parameterized by the same $\theta$, but with tag set $Y'$, transition
function $g'_\theta$, and emission function $h'_\theta$. This
auxiliary CRF describes the distribution $P'_\theta(\bm{y'}\mid\x)$
(with $\bm{y'} \in Y'^*$), which we will be able to interpret as
$\Pc$.  The construction is as follows:
\begin{gather}
Y' = E \\
g'_{\theta}((q, a, r),(q', a', r')) =
\begin{cases}
    g_{\theta}(a, a') & \text{if } r = q'\\
    -\infty & \text{otherwise} \\
\end{cases} \\
h'_{\theta}(\x, (q, a, r), i) = \begin{cases}
    -\infty & \text{if } i = 1, q \neq q_1\\
    -\infty & \text{if } i = t, r \not\in F\\
    h_{\theta}(\x, a, i) & \text{otherwise.}\\
    \end{cases} 
\end{gather}
This means that the tags of our new CRF are the edges of $M$, the
transition function assigns zero probability to transitions between
edges which do not pass through a shared NFA state, and the emission
function assigns zero probability to tag sequences which do not begin
at the starting state or end at an accepting state.  Apart from these
constraints, the transition and emission functions depend only on edge
labels, and not on the edges themselves, and agree with the standard
CRF's $g_\theta$ and $h_\theta$ when
no constraints are violated.

As $M$ is unambiguous, every tag sequence $\y$ corresponds
to a single path through $M$, representable as an edge sequence
$\bm{\pi} = \langle\pi_1, \pi_2, ..., \pi_t\rangle, \pi_i \in E$.  Since this path is a
tag sequence for our auxiliary CRF, we can directly calculate the
auxiliary CRF's $P'_\theta(\bm{\pi}\mid\x)$.  From the construction of
$g'_\theta$ and $h'_\theta$, this must be equal to $\Pc$, as it is
proportional to $\Pu$ for $\y \in \Lang$, and zero (or, more
correctly, undefined) otherwise.  Figure~\ref{fig:construction}
illustrates this construction with a concrete example.

\subsection{Time and space efficiency}
\label{sec:tagmin}
As the Viterbi and forward algorithms are quadratic in $|Y|$,
very large tag sets can lead to performance issues, possibly making
training or inference intractible in extreme cases. Thus, we would like
to characterize under which conditions a RegCCRF can be used tractibly,
and identify techniques for improving performance.
As $Y$ corresponds to the edges of $M$, we would like to select our
unambiguous automaton $M$ to have as few edges as possible. For arbitrary languages,
this problem is NP-complete \citep{jiang1991minimal}, and, assuming
$\text{P} \neq \text{NP}$, is not even efficiently approximable
\citep{gruber2007inapproximability}. Nonetheless, for many common
classes of languages, there exist approaches to obtain a tractably
small automaton.

One straightforward method is to construct $M$ directly from a short
unambiguous regular expression. \cite{10.1007/3-540-55210-3_182} present
a simple algorithm for constructing an unambiguous automaton from an
unambiguous regular expression, with $|Q|$ linear in the length of the
expression.  Using this method to construct $M$, the time- and
space-complexity of Viterbi are polynomial in the length of our
regular expression, with a worst-case of quartic complexity
when the connectivity graph of $M$ is dense.

For many other tasks, a reasonable approach is to leverage domain
knowledge about the constraints to manually construct a small
unambiguous automaton. For example, if the constraints require that a
particular label occurs exactly $n$ times in the output sequence, an
automaton could be constructed manually to count ocurrences of that
label. Multiple constraints of this type can then be composed via 
automaton union and intersection.

Without making changes to $M$, we can also reduce the size of $|Y|$ by
adjusting our construction.
Instead of making each edge of $M$ a tag, we can adopt equivalence
classes of edges.  Reminiscent of standard NFA minimization, we define
$(q, a, r) \sim (q', a', r') \leftrightarrow (q, a) = (q', a')$. When
constructing our CRF, whenever a transition would have been allowed
between two edges, we allow a transition between their corresponding
equivalence classes.  We do the same to determine which classes are
allowed as initial or final tags.  As each equivalence class
corresponds (non-uniquely) to a single symbol $a$, we can translate
between tag sequences and strings of $\mathcal{L}$ just as before.

\subsection{Interpretation as a weighted finite-state
transducer}
\label{sec:fsts}
While we present our model as a variation of a standard CRF which
enforces regular-language constraints, an alternate characterization is
as a weighted finite-state transducer with the transducer topology
and weight parameterization chosen so as to yield the distribution $\Pc$.
In order to accommodate CRF transition weights, such an approach involves
weight-learning in an auxiliary automaton whose edges correspond to
edge-pairs in $M$ -- we give a full construction in
Appendix~\ref{sec:appendix_fst_construction}.

This interpretation enables direct comparison to studies on weight
learning in finite-state transducers, such as
\cite{rastogi-etal-2016-weighting}. While RegCCRFs can be viewed as
special cases of neural-weighted FSTs, they inherit a number of useful
properties from CRFs not possessed by neural-weighted automata in
general. Firstly, as $|\y|$ is necessarily equal to $|\x|$, the
partition function $\sum_{\bm{y} \in \Lang}\Pc$ is guaranteed to be
finite, and $\Pc$ is a well-defined probability distribution for all
$\theta$, which is not true for weighted transducers in general, which
may admit paths with unbounded lengths and weights. Secondly, as $M$
is assumed to be unambiguous, string probabilities correspond exactly
to path probabilities, allowing for exact MAP inference with the
Viterbi algorithm. In contrast, finding the most probable string in
the highly ambiguous automata used when learning edge weights for an
unknown language is NP-Hard \citep{1999PaReL..20..813C}, necessitating
approximation methods such as crunching
\citep{may-knight-2006-better}. Finally, as each RegCCRF can be
expressed as a CRF with a particular parameterization, the convexity
guarantees of standard CRFs carry over, in that the loss is convex
with respect to emission and transition scores. In
contrast, training losses in general weighted finite-state transducers
are usually nonconvex \citep{rastogi-etal-2016-weighting}.

\section{Constrained training vs. constrained decoding}
\label{sec:constrained_decoding}
Our construction suggests two possible use cases for a RegCCRF:
\textit{constrained decoding}, where a CRF is trained unconstrained, and
the learned weights are then used in a RegCCRF at decoding time, and
\textit{constrained training}, where a RegCCRF is both trained and decoded
with constraints.
In this section, we compare between these two approaches and a standard,
\textit{unconstrained CRF}.
We assume a machine learning
setting where we have access to samples from some data distribution
$\widetilde{P}(\x, \y)$, with each $\x \in X^*$, and each $\y$ of
matching length in some regular language $\Lang \subseteq{\Sigma^*}$.
We wish to model the conditional distribution
$\widetilde{P}(\y\mid\x)$ with either a CRF or a RegCCRF, by way of maximizing
the model's (log) likelihood given the data distribution.

The unconstrained CRF corresponds to a CRF that has been trained,
without constraints, on data from $\widetilde{P}(\x, \y)$, and is used
directly for inference: It makes no use of the language $\Lang$.  The
model's output distribution is $\Pu[\thetau]$, with parameter vector
$\thetau$ minimizing the NLL objective:
\begin{equation}
\thetau = \argmin_\theta \E{- \ln \Pu} \label{eq:thetau}
\end{equation}
In constrained decoding, a CRF is trained unconstrained, 
but its weights are used in a RegCCRF at decoding time. The output distribution
of such a model is
$\Pc[\thetau]$.
It is parameterized by the same
parameter vector $\theta_u$ as the unconstrained CRF, as the training procedure is
identical, but the output distribution is conditioned on $\y \in \Lang$.

Constrained training involves directly optimizing a
RegCCRF's output distribution, avoiding any asymmetry between training and decoding time.
In this case, the output distribution of the model is $\Pc[\thetac]$,
where
\begin{equation}
\theta_c = \argmin_\theta \E{-\ln \Pc}
\end{equation}
is the parameter vector which minimizes the NLL of the RegCCRF's
constrained distribution.

These three approaches form a hierarchy in terms of their ability to
match the data distribution: 
$L_\text{unconstrained} \geq L_\text{constrained decoding} \geq
L_\text{constrained training}$,
with each $L$ corresponding to the negative log-likelihood assigned by each model to the data;
see Appendix~\ref{sec:appendix_proof} for a proof. This suggests we should prefer the constrained training regimen.

\section{Synthetic Data Experiments}

While  constrained training cannot underperform
constrained decoding, the conditions where it is strictly better
depend on exactly how the transition and emission functions are
parameterized, and are not easily stated in general terms.  We now
empirically show two simple experiments on synthetic data where the two are not equivalent.

The procedure is similar for both experiments. We specify a regular language $\Lang$, an observation
alphabet $X$, and a joint data distribution $\widetilde{P}(\x, \y)$
over observation sequences in $X^*$ and label sequences in $\Lang$.
We then train two models, one with a RegCCRF, parameterized by
$\theta_{c}$, and one with an unconstrained CRF, parameterized by
$\theta_{u}$.  For each model, we initialize parameters randomly,
then use stochastic gradient descent to minimize NLL with $\widetilde{P}$.
We directly generate samples from
$\widetilde{P}$ to use as training data.  After optimizing
$\theta_{c}$ and $\theta_{u}$, construct a RegCCRF with $\theta_u$
for use as a constrained-decoding model, and we compare the constrained-training
distribution $\Pc[\theta_c]$ with the constrained-decoding
distribution $\Pc[\theta_u]$.

We use a simple architecture for our models, with both the transition functions
$g_\theta$ and emission functions $h_\theta$ represented as parameter matrices.
We list training hyperparameters in Appendix~\ref{sec:appendix_experimental}.

\subsection{Arbitrarily large differences in likelihood}

\begin{figure}
\resizebox{\textwidth}{!}{
\begin{tikzpicture}

\definecolor{color0}{rgb}{1,0.498039215686275,0.0549019607843137}
\definecolor{color1}{rgb}{0.172549019607843,0.627450980392157,0.172549019607843}

\begin{axis}[
legend cell align={left},
legend style={fill opacity=0.8, draw opacity=1, text opacity=1, at={(0.97,0.03)}, anchor=south east, draw=white!80!black},
tick align=outside,
tick pos=left,
x grid style={white!69.0196078431373!black},
xlabel={\(\displaystyle k\)},
xmin=1, xmax=20,
xtick style={color=black},
y grid style={white!69.0196078431373!black},
ylabel={\(\displaystyle P((\texttt{ac})^k)\)},
ymin=0, ymax=1.1,
ytick style={color=black}
]
\addplot [thick, color0, mark=square*, mark size=1.5, mark options={solid}]
table {%
1 0.749697519854268
2 0.901400564937665
3 0.964572896425071
4 0.988006613479835
5 0.996057672439138
6 0.998814332610071
7 0.999527089396549
8 0.999847423750315
9 0.999931337806575
10 0.999977112078339
11 1
12 0.999984741327352
13 1.00002288844553
14 1.00002288844553
15 1.00002288844553
16 0.999992370634572
17 1
18 0.999984741327352
19 1.00004577741494
20 1
};
\addlegendentry{Constrained decoding}
\addplot [thick, color1, mark=*, mark size=1.5, mark options={solid}]
table {%
1 0.748036304522599
2 0.7499128676391
3 0.748813492907948
4 0.751114940143684
5 0.749803096673471
6 0.758247538301655
7 0.753148362534024
8 0.746385265414588
9 0.749342667297559
10 0.761385444520194
11 0.742852228733231
12 0.736359235531681
13 0.748202853407461
14 0.756888972613914
15 0.773991434250918
16 0.752891808480769
17 0.739626468160279
18 0.744932012475684
19 0.743236656377097
20 0.754834023008278
};
\addlegendentry{Constrained training}
\addplot [ultra thick, red, opacity=0.5, dotted]
table {%
1 0.75
21 0.75
};
\addlegendentry{Data distribution}
\end{axis}

\end{tikzpicture}
\begin{tikzpicture}

\definecolor{color0}{rgb}{1,0.498039215686275,0.0549019607843137}
\definecolor{color1}{rgb}{0.172549019607843,0.627450980392157,0.172549019607843}

\begin{axis}[
legend cell align={left},
legend style={fill opacity=0.8, draw opacity=1, text opacity=1, at={(0.03,0.97)}, anchor=north west, draw=white!80!black},
tick align=outside,
tick pos=left,
x grid style={white!69.0196078431373!black},
xlabel={\(\displaystyle k\)},
xmin=1, xmax=20,
xtick style={color=black},
y grid style={white!69.0196078431373!black},
ylabel={Negative log-likelihood},
ymin=0.320362550020218, ymax=5.64375938773155,
ytick style={color=black}
]
\addplot [thick, color0, mark=square*, mark size=1.5, mark options={solid}]
table {%
1 0.56233537197113
2 0.657027721405029
3 0.862116575241089
4 1.11489772796631
5 1.38703536987305
6 1.68456220626831
7 1.91788291931152
8 2.20881938934326
9 2.45114612579346
10 2.82891368865967
11 3.01506423950195
12 3.27549362182617
13 3.54771041870117
14 3.81138610839844
15 4.14768600463867
16 4.35474586486816
17 4.50785446166992
18 4.9239501953125
19 5.03129959106445
20 5.40178680419922
};
\addlegendentry{Constrained decoding}
\addplot [thick, color1, mark=*, mark size=1.5, mark options={solid}]
table {%
1 0.562345385551453
2 0.562335133552551
3 0.562338888645172
4 0.562338471412659
5 0.562335252761841
6 0.562519252300262
7 0.562361836433411
8 0.562369823455811
9 0.562336146831512
10 0.562688112258911
11 0.562469720840454
12 0.562820076942444
13 0.562344074249268
14 0.562463045120239
15 0.563942193984985
16 0.562358796596527
17 0.562616944313049
18 0.562404453754425
19 0.562455832958221
20 0.562398612499237
};
\addlegendentry{Constrained training}
\addplot [ultra thick, red, opacity=0.5, dotted]
table {%
1 0.562335144618808
21 0.562335144618808
};
\addlegendentry{Data distribution entropy}
\end{axis}

\end{tikzpicture}
}
\caption{Model output probabilities, and NLL losses,
	plotted against sequence length $k$.  As $k$
  increases, constrained decoding becomes a progressively worse
  approximation for the data distribution, while constrained training
  is consistently able to match the data distribution.}
\label{fig:arbitrary}
\end{figure}
We would like to demonstrate that, when comparing constrained training
to constrained decoding in terms of likelihood, constrained training can outperform constrained decoding
by an arbitrary margin.  We choose
$\Lang = (\texttt{ac})^*\mid(\texttt{bc})^*$ to make conditional
independence particularly relevant -- as every even-indexed label is
\texttt{c}, an unconstrained CRF must model odd-indexed labels
independently, while a constrained CRF can use its constraints to
account for nonlocal dependencies.  For simplicity, we hold the input
sequence constant, with $X = \{\texttt{o}\}$.

Our approach is to construct sequences of various lengths. For $k \in \mathbb{N}$, we let our data distribution be
\begin{equation}
\widetilde{P}(\texttt{o}^{2k}, (\texttt{ac})^k) = \frac{3}{4} \text{
  and }
\widetilde{P}(\texttt{o}^{2k}, (\texttt{bc})^k) = \frac{1}{4}.
\end{equation}
As the marginal distributions for odd-indexed characters are not
independent, an unconstrained CRF cannot exactly
represent the distribution $\widetilde{P}$.
We train and evaluate individual models for each sequence length
$k$.
Figure~\ref{fig:arbitrary} plots model probabilities and
NLL losses for various $k$.  We see that, regardless of $k$,
$\Pc[\theta_c]$ is able to match $\widetilde{P}(\y\mid\x)$ almost
exactly, with only small deviations due to sampling noise in SGD.  On
the other hand, as sequence length increases, $\Pc[\theta_u]$ becomes
progressively ``lop-sided'', assigning almost all of its probability
mass to the string $(\texttt{ac})^k$.  This behavior is reflected in the
models' likelihoods -- constrained
training stays at near-constant likelihood for all $k$, while the
negative log-likelihood of constrained decoding grows linearly with $k$.

\subsection{Differences in MAP inference}
\label{sec:map}
We show here that constrained training and constrained decoding can disagree
about which label sequence they deem most likely.
Furthermore, in this case, MAP inference agrees with the data distribution's
mode for constrained training, but not for constrained decoding.
To do this, we construct a fixed-length output language
$\Lang = \texttt{acd}\mid\texttt{bcd}\mid\texttt{bce}$, where an
unconstrained CRF is limited by the Markov property to predict $\y$'s prefix
and suffix independently, and choose a data distribution which violates
this independence assumption.
We select our data distribution,
\begin{equation}
\widetilde{P}(\texttt{ooo}, \texttt{acd}) = 0.4 \text{ and }
\widetilde{P}(\texttt{ooo}, \texttt{bcd}) = 0.3 \text{ and }
\widetilde{P}(\texttt{ooo}, \texttt{bce}) = 0.3,
\end{equation}
to be close to uniform, but with one label sequence holding
the slight majority, and we ensure that the majority label sequence is \textit{not}
the label sequence with both the majority prefix and the majority suffix (i.e. \texttt{bcd}).
As before, we hold the observation sequence as a constant ($\texttt{ooo}$).
We train a constrained and an unconstrained CRF to convergence, and compare $\Pc[\thetau]$
to $\Pc[\thetac]$.

\begin{table}
\centering
\caption{Output distributions for constrained decoding ($\Pc[\thetau]$)
and constrained training ($\Pc[\thetac]$), compared to the data distribution $\widetilde{P}(\y\mid\x)$.
Constrained decoding cannot learn the data distribution exactly, and yields a
mode which disagrees with that of the
data distribution.}
\label{tab:map}
\begin{tabular}{lccc}
\toprule
$\y$ & $\widetilde{P}(\y\mid\x)$ & $\Pc[\thetau]$ & $\Pc[\thetac]$ \\
\cmidrule(r){1-1}\cmidrule(rl){2-2}\cmidrule(rl){3-3}\cmidrule(l){4-4}
\texttt{acd} & \textbf{0.4} & 0.32 & \textbf{0.40}\\
\texttt{bcd} & 0.3 & \textbf{0.48} & 0.30\\
\texttt{bce} & 0.3 & 0.20 & 0.30\\
\bottomrule
\end{tabular}
\end{table}

Table~\ref{tab:map} shows $\Pc[\thetau]$ and $\Pc[\thetac]$ as they compare to
$\widetilde{P}(\y\mid\x)$.
We find that, while the mode of $\widetilde{P}(\y\mid\x)$ is \texttt{acd},
with probability of 0.4,
the mode of constrained decoding distribution $\Pc[\thetau]$ is \texttt{bcd},
the string with the majority prefix and the majority suffix, to which the
model assigns a probability of 0.48.  Conversely, the constrained
training distribution $\Pc[\thetac]$ matches the data distribution almost
exactly, and predicts the correct mode.

\section{Real-world data experiment: semantic role labeling}

\paragraph{Task.} As a final experiment, we apply our RegCCRF to the
NLP task of semantic role labeling (SRL) in the popular PropBank
framework \citep{palmer-etal-2005-proposition}. In line with previous
work, we adopt the \textit{known-predicate setting}, where events are
given and the task is to mark token spans as (types of) event
participants. PropBank assumes 7 semantic \textit{core roles} (ARG0
through ARG5 plus ARGA) plus 21 \textit{non-core roles} for modifiers
such as times or locations. For example, in [$_{\text{ARG0}}$
\textit{Peter}] \textbf{saw} [$_{\text{ARG1}}$ \textit{Paul}]
[$_{\text{ARGM-TMP}}\, \textit{yesterday}]$, the argument labels
inform us who does the seeing (ARG0), who is seen (ARG1), and when the
event took place (ARGM-TMP). In addition, role spans may be labeled as
\textit{continuations} of previous role spans, or as
\textit{references} to another role span in the sentence.
SRL can be framed naturally as a sequence labeling task
\citep{he2017deep}. However, the task comes with a number of hard
constraints that are not automatically satisfied by standard CRFs,
namely: (1) Each core role may occur at most once per event; (2) for
continuations, the span type must occur previously in the sentence;
(3) for references, the span type must occur elsewhere in the sentence.


\paragraph{Data.} In line with previous work
\citep{ouchi-etal-2018-span}, we work with the OntoNotes corpus as
used in the CoNLL 2012 shared task\footnote{
As downloaded from \url{https://catalog.ldc.upenn.edu/LDC2013T19}, and
preprocessed according to \url{https://cemantix.org/data/ontonotes.html}
}
 \citep{ontonotes, pradhan-etal-2012-conll},
whose training set comprises 66 roles (7 core roles, 21 non-core roles,
19 continuation types, and 19 reference types).

\paragraph{RegCCRF Models.} 
To encode the three constraints listed above in a
RegCCRF, we define a regular language describing valid BIO
sequences \citep{ramshaw1999text} over the 66 roles. A minimal
unambiguous NFA for this language has more than
$2^2\cdot3^{19}$ states, which is too large to run the Viterbi
algorithm on our hardware.  However, as many labels are very rare,
we can shrink our automaton by discarding them at the cost of
imperfect recall. We achieve further reductions in size by ignoring
constraints on reference roles, treating them identically to non-core
roles. Our final automaton recognizes 5 core role types (ARG0 through
ARG4), 17 non-core / reference roles, and one continuation role type
(for ARG1). This automaton has 672 states, yielding a RegCCRF with
2592 tags. A description of our procedure for constructing this automaton
can be found in Appendix~\ref{sec:appendix_nfa_construction}.

Our model architecture is given by this RegCCRF, with emission scores
provided by a linear projection of the output of a pretrained RoBERTa
network \cite{liu2019roberta}.  In order to provide the model with
event information, the given predicates are prefixed by a special
reserved token in the input sequence. RoBERTa parameters are
fine-tuned during the learning of transition scores and projection
weights. We perform experiments with both constrained training and
constrained decoding settings -- we will refer to these as \ct and \cd
respectively. A full description of the training procedure, including training times, is provided in Appendix~\ref{sec:appendix_experimental}.
As RegCCRF loss is only finite for label sequences in $\Lang$, we must
ensure that our training data do not violate our constraints. We
discard some roles, as described above, by simply removing the
offending labels from the training data. In six cases, training
instances directly conflict with the constraints specified --- all
cases involve continuation roles missing a valid preceding role.  We
discard these instances for \ct.

\paragraph{CRF Baselines.} As baseline models, we use the same
architecture, but with a standard CRF replacing the RegCCRF.  Since
we are not limited by GPU memory for CRFs, we are optionally able to include all
role types present in the training set, using the complete training
set.  We present two CRF baseline models: \textit{CRF-full}, which is trained on
all role-types from the training set, and \textit{CRF-reduced}, which includes the
same subset of roles as the RegCCRF models. For \textit{CRF-reduced}, we use the
same learned weights as for CD, but we decode without constraints.

\paragraph{Results and Analysis.}

\newcommand{\ranktext}[1]{\textsuperscript{(#1)}}
\newcommand{\rank}[1]{^{(#1)}}
\begin{table}
\caption[
  Results from our experiments, along with selected
  reported results from recent literature. 
  ]{
  Results from our experiments (averaged over twelve trials), along with selected
  reported results from recent literature. We rank of our models by
  precision, recall, and $F_1$ score -- there exists a significant difference
  between two comparable values if and only if their rankings differ by
  one or more.
  Statistical significance is reported at $p<0.05$ (two-tailed), as measured
  by a permutation test.
  }
\centering
\sisetup{input-symbols = {()},
 group-digits  = false,
 table-sign-mantissa,
 table-space-text-post = *,
 parse-numbers=false,
 detect-weight=true,
 detect-shape=true,
 detect-mode=true
 }

  \resizebox{\textwidth}{!}{
  \begin{tabular}{llSSS}
  \toprule
&   Model& {Precision\ranktext{rank}} & {Recall\ranktext{rank}} & {$F_1$\ranktext{rank}}\\
  \cmidrule(lr){2-2}\cmidrule(lr){3-3}\cmidrule(lr){4-4}\cmidrule(l){5-5}
\multirow{3}{*}{\parbox{15mm}{Our \\ models}}
&   CRF (\textit{\full}) & 86.89 \rank{2} & 87.81\rank{1} & 87.35\rank{2} \\
&   CRF (\textit{\red}) & 86.92 \rank{2} & 87.35\rank{2} & 87.13\rank{3} \\
&   RegCCRF (\textit{\cd}) & 87.05 \rank{1.5} & 87.38\rank{2} & 87.21\rank{2.5} \\
&   RegCCRF (\textit{\ct}) & 87.31 \rank{1} & 87.76\rank{1} & 87.53\rank{1} \\
  \cmidrule(r){1-2}\cmidrule(lr){3-3}\cmidrule(lr){4-4}\cmidrule(l){5-5}
\multirow{4}{*}{\parbox{15mm}{Results from\\ literature}}
&  \cite{he2017deep} & {---} & {---} & 85.5\\
&    \cite{ouchi-etal-2018-span} & 87.1 & 85.3 &  86.2\\
&   \cite{ouchi-etal-2018-span} \scriptsize{Ensemble} & 88.5 & 85.5 & 87.0\\
& \cite{Li_He_Zhao_Zhang_Zhang_Zhou_Zhou_2019} &  85.7 & 86.3 & 86.0 \\
  \bottomrule
  \end{tabular}
  }
  \label{tab:srl_results}
\end{table}

We evaluate our models on the
evaluation partition, and measure performance using $F_1$ score for exact span
matches.
For comparability with prior work, we use the evaluation script\footnote{As 
available from \url{https://www.cs.upc.edu/~srlconll/soft.html}.}
for the CoNLL-2005 shared task \citep{carreras-marquez-2005-introduction}.
These results, averaged over twelve trials, are presented in Table~\ref{tab:srl_results}.

All of our models outperformed the ensemble model of
\citep{ouchi-etal-2018-span}, which represents the state of the art for
this task as of publication of this work.
We ascribe this improvement over
the existing literature to our use of RoBERTa -- prior work in SRL
relies on ELMo \citep{peters-etal-2018-deep}, which tends to
underperform transformer-based models on downstream tasks
\citep{devlin-etal-2019-bert}.

Of our models, \ct significantly%
\footnote{All significance results are at the $p<0.05$ level (two-tailed), as
  measured by a permutation test over the twelve trials of each model.} outperforms the others
in $F_1$ score and yields a new SOTA for SRL on OntoNotes, in line with expectations from theoretical analysis and
on synthetic data.
For our unconstrained models, \textit{CRF-full} and \textit{CRF-reduced}, the constraints
specified in our automaton are violated in 0.81\% and 0.84\% of all
output sequences respectively.\footnote{For \textit{CRF-full}, we only count
  violations of constraints for those roles that our automaton accounts
  for.}

Among our four models, we see interesting trade-offs between precision
and recall.
For precision, while not all comparisons reach statistical significance,
models with constraints seem to outperform those without.
This is not surprising at all: the purpose of constraints is to prevent
models from making predictions  we know to be false a priori, and so we
should expect constrained models to be more precise overall.

For recall, we observe two clusters: \full and \ct both perform a bit
better, while \red and \cd both perform a bit worse.
As \full is the only model using the full tag set, and thus the only one
capable of predicting rare role types, its high recall is to be
expected.
Our other three models show an interesting pattern with regards to
recall: \red and \cd perform about at parity, with \ct showing
significant improvements.
This behavior turns out to be quite intuitive: \red and \cd were trained
identically, and only differ in their decoding procedure.
The decoding-time constraints of \cd largely work to prevent spurious
predictions, and so we shouldn't expect these models to differ much in
the number of true roles they predict.
On the other hand, since \ct is trained with constraints, it can learn
to be less ``cautious,'' with its predictions, as its constraints will
automatically prevent many false positives.
Thus, at evaluation time, \ct finds more roles that were avoided by \red
and \cd.

\section{Conclusion and Future Work}
We have presented a method to constrain the output of CRFs to a
regular language. Our construction permits constraints to be used at
training or prediction time; both theoretically and empirically,
training-time constraining better captures the data
distribution. Conceptually, our approach constitutes a novel bridge
between constrained CRFs and neural-weighted FSTs.

Future work could target enhancing the model's expressibility, either
by allowing constraints to depend explicitly on the input as regular
relations, or by
investigating non-binary constraints, i.e., regular language-based
constraints with learnable weights. Additionally, regular language
induction (e.g. \cite{regex_induction1, regex_induction2}) could be
used to learn languages automatically, reducing manual specification
and identifying non-obvious constraints.  Another avenue for
continuing research lies in identifying further applications for
RegCCRFs. The NLP task of relation extraction could be a fruitful
target -- RegCCRFs offer a mechanism to make the proposal of a
relation conditional on the presence of the right number and type of
arguments. While our construction cannot be lifted directly to
context-free languages due to the unbounded state space of the
corresponding pushdown automata, context-free language can be
approximated by regular languages \citep{mohri2001regular}. On this
basis, for example, a RegCCRF backed by a regular language describing
trees of a limited depth could also be applied to tasks with
context-free constraints.

To encourage the use of RegCCRFs, we provide an implementation as a
Python library under the Apache 2.0 license which can be used as a
drop-in replacement for standard CRFs in PyTorch.\footnote{
Available at \url{www.ims.uni-stuttgart.de/en/research/resources/tools/regccrf/}}

\section*{Acknowledgments}
This work is supported by IBM Research AI
through the IBM AI Horizons Network.

\section*{Reproducibility statement}
To ensure reproducibility, we have released all
code for the RegCCRF model as an open-source Python 3 library under the
Apache 2.0 license, which is included in the supplementary
materials. Additionally, we include Python scripts for reproducing all
experiments presented in the paper, detailed descriptions of our
datasets and preprocessing steps, and training logs. Furthermore,
Appendix~\ref{sec:appendix_experimental} lists all model
hyperparameters, details the preprocessing steps taken for our
experiments, and specifies the hardware used for our experiments along
with average training and inference times for each experiment.

\section*{Ethics statement}
The research in this paper
is fundamental in the sense that it enables machine learning models to
better represent data and limit the search space at inference and
learning time. It therefore does not in and of itself represent additional
ethical risks on top of the previous work we build upon.
\bibliographystyle{iclr2022_conference}
\bibliography{lit}

\begin{appendices}

\section{Experimental Design}
\label{sec:appendix_experimental}

\begin{table}
\caption{
\label{tab:hyper} Summary of hyperparameters for our models and experiments.
}
\centering
\begin{tabular}{llr}
\toprule
\multirow{1}{3cm}{\textbf{CRFs}} & Transition score initialization & $\mathcal{N}(0, 0.1)$ \\
\midrule
\multirow{6}{3cm}{\textbf{Synthetic data experiments}} &
Emission score initialization & PyTorch default \\
& Optimizer & SGD \\
& Training iterations & 5000 \\
& Batch size & 50 \\
& Initial learning rate & 1.0\\
& Learning rate decay & 10\% every 100 steps \\
\midrule
\multirow{6}{3cm}{\textbf{SRL experiments}} &
RoBERTa weights & \texttt{roberta-base} \\
& Projection weight and bias initialization & PyTorch default \\
& Optimizer & Adam \\
& Learning rate & $10^{-5}$\\
& Batch size & 2 \\
& Gradient accumulation & 4 batches \\
\bottomrule
\end{tabular}
\end{table}

This appendix details the training procedures and hyperparameter choices for
our experiments.
These are summarized in Table~\ref{tab:hyper}.
Full code for all experiments, along with training logs, are also included in
the supplementary materials.

\subsection{CRFs}
For all CRFs and RegCCRFs, transition potentials were initialized randomly
from a normal distribution with mean zero and standard deviation 0.1.
No CRFs or RegCCRFs employed special start- or end-transitions -- that is, we did not
insert any additional beginning-of-sequence or end-of-sequence tags for the
Viterbi or forward algorithms.

\subsection{Synthetic data experiments -- training procedure}
For both synthetic data experiments, the emission potentials were represented
explicitly for each position as trainable parameters -- since the observation
sequence was constant in all experiments, these did not depend on  $\bm{x}$.

Parameters were initialized randomly using PyTorch default initialization, and
optimized using stochastic gradient descent. To ensure fast convergence to a
stable distribution, we employed learning rate decay -- learning rate was initially
set to 1.0, and reduced by 10\% every 100 training steps.

We trained all models for a total of 5000 steps with a batch size of 50.
All models were trained on CPUs.
For the experiment described in Section~6.1, we trained separate models for each
$k$ -- the total training time for this experiment was approximately 35 minutes.
The experiment described in Section~6.2 completed training in approximately 30
seconds.

\subsection{Semantic role labeling -- training procedure}
In the semantic role labeling (SRL) experiments, we incorporated a pretrained RoBERTa
network \citep{liu2019roberta} -- the implementation and weights for this model
were obtained using the \texttt{roberta-base} model from the Hugging Face
transformers library \citep{wolf-etal-2020-transformers}. RoBERTa embeddings
were projected down to transmission scores using a linear layer with a bias --
projection weights and biases were initialized using the PyTorch default
initialization.

Input tokens were sub-tokenized using RoBERTa's tokenizer.
The marked predicate in each sentence was prefixed by a special \texttt{<unk>}
token.
During training, for efficiency reasons, we excluded all sentences with 120 or
more subtokens -- this amounted to 0.23\% of all training instances.
We nonetheless predicted on all instances, regardless of length.

We optimized models using the Adam optimizer \citep{adam} with a learning rate
of $10^{-5}$.
We fine-tune RoBERTa parameters and learn the projection and RegCCRF weights
jointly.
For performance reasons, batch size was set to 2, but we utilized gradient
accumulation over groups of 4 batches to simulate a batch size of 8.

We utilized early stopping to avoid overfitting.
Every 5000 training steps, we approximated our model's $F_1$ score against a
subset of the provided development partition, using a simplified reimplementation of the
official evaluation script.
Each time we exceeded the previous best $F_1$ score for a model, we saved all
model weights to disk.
After 50 such evaluations with no improvement, we terminated training, and used
the last saved copy of model weights for final evaluation.

We performed all SRL experiments on GeForce GTX 1080 Ti GPUs.
Each experiment used a single GPU.
Training took an average of 88 hours for RegCCRF models with constrained
training, 23 hours for RegCCRF with constrained decoding, and 24 hours for CRF
baseline models.  Inference on the complete test set took an average of
18 minutes 55 seconds for CT and CD, and an average of 55 seconds for
CRF-full and CRF-reduced. 
All training logs with timestamps are included in the
supplementary materials.

\section{Proof of constrained training inequality}
In this appendix, we prove the inequality presented in section \ref{sec:constrained_decoding}:
when compared by NLL against the data distribution, 
$L_\text{unconstrained} \geq L_\text{constrained decoding} \geq
L_\text{constrained training}$,
with each $L$ corresponding to that model's negative log-likelihood.
We first prove the left side of this inequality, comparing an
unconstrained CRF to constrained decoding, and then prove the right side,
comparing constrained decoding to constrained training. We use the
notation introduced in Section~\ref{sec:constrained_decoding}.
\label{sec:appendix_proof}
\begin{theorem}{For arbitrary $\theta$: $\E{- \ln \Pu} \geq \E{- \ln \Pc}$}
\end{theorem}
Here we compare the distributions $\Pu[\theta]$ and $\Pc[\theta]$.
We wish to demonstrate that $\Pu[\theta]$ can never achieve lower NLL
than $\Pc[\theta]$, and that the two distributions
achieve identical NLL only when $\Pu[\theta] = \Pc[\theta]$ i.e. when constraints have no effect.
Of note, this proof is valid for \textit{all} parameterizations $\theta$,
and not just for $\thetau$.
\begin{proof}
Since every $\y$ in $\widetilde{P}$ is in $\Lang$,
\begin{equation}
\Pc = \alpha \cdot \Pu,
\end{equation}
with $\alpha \geq 1$.
Thus, the NLL of the regular-constrained CRF is
\begin{equation}
\E{- \ln \Pc} = \E{- \ln \Pu} - \ln \alpha.
\end{equation}
This differs from the NLL of the unconstrained CRF only by the term $- \ln \alpha$.
As $\alpha \geq 1$, the regular-constrained CRF's NLL is less than or equal to that of the unconstrained CRF,
with equality only when $\alpha = 1$ and therefore $\Pu = \Pc$. 
\end{proof}

\begin{theorem} $\E{- \ln \Pc[\thetau]} \geq \E{- \ln \Pc[\thetac]}$
\end{theorem}
In this case, we compare the distributions $\Pc[\thetau]$ and $\Pc[\thetac]$.
We will demonstrate that the former cannot achieve a lower NLL against the
data distribution than the latter.
\begin{proof}
This follows directly from our definitions, as we
define $\thetac$ to minimize the NLL of \mbox{$\Pc$} against the data
distribution.  Thus, $\Pc[\thetau]$ could never yield a lower NLL
than $\Pc[\thetac]$, as that would contradict our definition of $\thetac$.
\end{proof}

\section{Construction as weighted FST}
\label{sec:appendix_fst_construction}
In this appendix, we present a construction of the RegCCRF as a weighted
finite-state transducer with weight sharing. We do this by first
specifying the transducer topology used, and then specifying how edge
weights are parameterized in terms of $\theta$. The resulting transducer
yields an identical distribution to that of the CRF-based construction,
$\Pc$.
\subsection{Transducer topology}
Starting from $\Lang$, we define $\biLang$ to be the regular language of
bigram sequences for the strings in $\Lang$, i.e.,
\begin{equation}
\biLang = \left \{ \left\langle (s_1, s_2), (s_2, s_3), ..., (s_{|\bm{s}-1|}, s_{|\bm{s}|}), (s_{|\bm{s}|}, \$) \right\rangle \mid \bm{s} \in \Lang \right \},
\end{equation}
with $\$$ acting as a end-of-string symbol.
We let $\biM$ be an unambiguous FSA for the language $\biLang$, and choose
to interpret this automaton as a finite-state transducer by stipulating
that all edges should accept any symbol in the input language (but only
one symbol per transition, and without allowing epsilon transitions).
This unweighted transducer will be used as the topology for our weighted
finite-state transducer.
\subsection{Edge weights}
In line with \cite{rastogi-etal-2016-weighting}, we would like to assign
weights to the edges of our transducer $\biM$ with a neural network. In
order to obtain the same distribution as from our CRF-based construction,
these weights must be parameterized in terms of our transition function
$g_{\theta}$ and emission function $h_{\theta}$.
For each edge in $\biM$, the weight depends only on the emitted bigram,
the input sequence, and the index of the current input
symbol -- the weight does not depend on the FST states. 
For a symbol bigram $(a, b)$, input sequence $\bm{x}$, and index $i$,
the edge weight is equal to
\begin{equation}
W_{a,b} = \begin{cases}
g_\theta(a, b) + h_\theta(\x, a, i) & b \neq \$ \\
h_\theta(\bm{x}, a, i) & \text{otherwise}
\end{cases}.
\end{equation}

Each string in $\Lang$ corresponds bijectively to exactly one bigram
sequence in $\biLang$, which corresponds bijectively to exactly one
accepting path in $\biM$ -- this path's weight (in the Log semiring)
is equal to the unscaled probability produced by our CRF construction,
and so the weighted FST, interpreted as a probability distribution,
yields the distribution $\Pc$.

\section{Automaton construction for semantic role labeling}
\label{sec:appendix_nfa_construction}
\newcommand{\rcore}{\mathcal{R}_{\text{core}}}
\newcommand{\rnoncore}{\mathcal{R}_{\text{noncore}}}
\newcommand{\rcont}{\mathcal{R}_{\text{continuation}}}

In this appendix, we describe how we generate the automaton
architecture for our semantic
role labeling experiments. While our experiments used 5 core-roles, 17
non-core roles, and one continuation role, we discuss here a generalized setting
with arbitrary sets of core, noncore, and continuations of core roles.

Algorithm~\ref{alg:nfa} provides pseudocode for our construction. The
core idea is to use subsets of core roles as NFA states, so that we can
keep track of which core roles have already ocurred. Additional states
are used in order to ensure all strings are valid BIO sequences.

\begin{algorithm}[H]
	\KwData{Sets $\rcore$, $\rnoncore$, and $\rcont$, of core, noncore, and continuation roles, respectively}
	\KwResult{A finite-state automaton $M = (\Sigma, Q, q_1, F, E)$, parameterized as described in Section~\ref{sec:prelim}}
	$\Sigma \leftarrow \{\textsc{Outside}\} \cup (\{\textsc{Begin}, \textsc{Inside}\} \times (\rcore\cup\rnoncore\cup\rcont))$\;
	$Q \leftarrow \varnothing$\;
	$q_1 \leftarrow \varnothing$\;
	$F \leftarrow \varnothing$\;
	$E \leftarrow \varnothing$\;
	\For{$p \in 2^{\rcore}$} {
		$Q \leftarrow Q \cup \{p\}$\;
		$F \leftarrow F \cup \{p\}$\;
		$E \leftarrow E \cup \{(p, \textsc{Outside}, p)\}$\;
		\For{$r \in \rnoncore$} {
			$s \leftarrow (r, p)$\;
			$Q \leftarrow Q \cup \{s\}$\;
			$E \leftarrow E \cup \{(p, (\textsc{Begin}, r), p),(p, (\textsc{Begin}, r), s)\}$\;
			$E \leftarrow E \cup \{(r, (\textsc{Inside}, r), s),(r, (\textsc{Inside}, r), p)\}$\;
		}
		\For{$r \in \rcont$} {
			\If{The core role corresponding to $r$ is in $p$} {
				$s \leftarrow (r, p)$\;
				$Q \leftarrow Q \cup \{s\}$\;
				$E \leftarrow E \cup \{(p, (\textsc{Begin}, r), p),(p, (\textsc{Begin}, r), s)\}$\;
				$E \leftarrow E \cup \{(r, (\textsc{Inside}, r), s),(r, (\textsc{Inside}, r), p)\}$\;
			}
		}
		\For{$r \in (\rcore \setminus p)$} {
			$s \leftarrow (r, p)$\;
			$t \leftarrow p \cup \{r\}$\;
			$Q \leftarrow Q \cup \{s\}$\;
			$E \leftarrow E \cup \{(p, (\textsc{Begin}, r), s),(p, (\textsc{Begin}, r), t)\}$\;
			$E \leftarrow E \cup \{(s, (\textsc{Inside}, r), s),(s, (\textsc{Inside}, r), t)\}$\;
		}
	}
	\Return{$(\Sigma, Q, q_1, F, E)$}
 \caption{\label{alg:nfa}Construction of an FSA from given sets of core, noncore, and continuation roles.
 To represent BIO labels, we use tuples of the form $(\textsc{Begin}, \texttt{<roleType>})$ for B labels,
 tuples of the form $(\textsc{Inside}, \texttt{<roleType>})$ for I labels, and the symbol $\textsc{Outside}$
 for the sole O label.
 }
\end{algorithm}

\end{appendices}

\end{document}